\documentclass[utf8]{frontiersSCNS} % for Science, Engineering and Humanities and Social Sciences articles
\usepackage{hyperref}
\usepackage{url,lineno,microtype,subcaption}
\usepackage[onehalfspacing]{setspace}

\def\keyFont{\fontsize{8}{11}\helveticabold }
\def\firstAuthorLast{}
\def\Authors{Roshan Gopalakrishnan\,$^{1}$, Yansong Chua\,$^{1,*}$ and Ashish Jith Sreejith Kumar\,$^{2}$}
\def\Address{$^{1}$Institute for Infocomm Research (I2R), ASTAR, Singapore \\
$^{2}$School of Electrical and Electronic Engineering, Nanyang Technological University, Singapore  }
\def\corrAuthor{Yansong Chua}

\def\corrEmail{chuays@i2r.a-star.edu.sg}

\begin{document}

\onecolumn
\firstpage{1}

%\title[Running Title]{Hardware-friendly Neural Network Architecture for Neuromorphic Computing}
\title{Hardware-friendly Neural Network Architecture for Neuromorphic Computing}  

\author[\firstAuthorLast ]{\Authors} %This field will be automatically populated
\address{} %This field will be automatically populated
\correspondance{} %This field will be automatically populated

\extraAuth{}% If there are more than 1 corresponding author, comment this line and uncomment the next one.
%\extraAuth{corresponding Author2 \\ Laboratory X2, Institute X2, Department X2, Organization X2, Street X2, City X2 , State XX2 (only USA, Canada and Australia), Zip Code2, X2 Country X2, email2@uni2.edu}

\maketitle

\begin{abstract}

%%% Leave the Abstract empty if your article does not require one, please see the Summary Table for full details.
%\section{}
%For full guidelines regarding your manuscript please refer to \href{http://www.frontiersin.org/about/AuthorGuidelines}{Author Guidelines}.

%As a primary goal, the abstract should render the general significance and conceptual advance of the work clearly accessible to a broad readership. References should not be cited in the abstract. Leave the Abstract empty if your article does not require one, please see \href{http://www.frontiersin.org/about/AuthorGuidelines#SummaryTable}{Summary Table} for details according to article type.

The hardware-software co-optimization of neural network architectures is becoming a major stream of research especially due to the emergence of commercial neuromorphic chips such as the IBM Truenorth and Intel Loihi. Development of specific neural network architectures in tandem with the design of the neuromorphic hardware considering the hardware constraints will make a huge impact in the complete system level application. In this paper, we study various neural network architectures and propose one that is hardware-friendly for a neuromorphic hardware with crossbar array of synapses. Considering the hardware constraints, we demonstrate how one may design the neuromorphic hardware so as to maximize classification accuracy in the trained network architecture, while concurrently, we choose a neural network architecture so as to maximize utilization in the neuromorphic cores. We also proposed
a framework for mapping a neural network onto a neuromorphic chip named as the Mapping and Debugging (MaD) framework. The MaD framework is designed to be generic in the sense that it is a Python wrapper which in principle can be integrated with any simulator tool for neuromorphic chips.

\tiny
 \keyFont{ \section{Keywords:} neuromorphic computing, neuromorphic chip, hardware constraints, deep learning, hardware-friendly, neural network, crossbar array, convolution, convolutional neural network} 
% All article types: you may provide up to 8 keywords; at least 5 are mandatory.
\end{abstract}

\section{Introduction}

%For Original Research Articles \citep{conference}, Clinical Trial Articles \citep{article}, and Technology Reports \citep{patent}, the introduction should be succinct, with no subheadings \citep{book}. For Case Reports the Introduction should include symptoms at presentation \citep{chapter}, physical exams and lab results \citep{dataset}.

Research in new architectures for convolutional neural networks (CNN) has progressed in various directions so as to fulfil different objectives: smaller deep neural networks \cite{MobileNet, SqueezeNet}, low precision neural networks \cite{BNN, XNORNET, DOREFA} and larger neural networks \cite{ResidualNET} etc. Among these new architectures, the smaller and low precision neural networks are more hardware friendly, in the sense that the entire network maybe mapped onto a neuromorphic chip. These neuromorphic chips are very power efficient as their computations are in spikes, which makes it a good candidate for low power applications in internet of things (IoT), unmanned aerial vehicles (UAVs), robotics and edge computing. 

\begin{figure}[h!]
\begin{center}
\includegraphics[width=18cm]{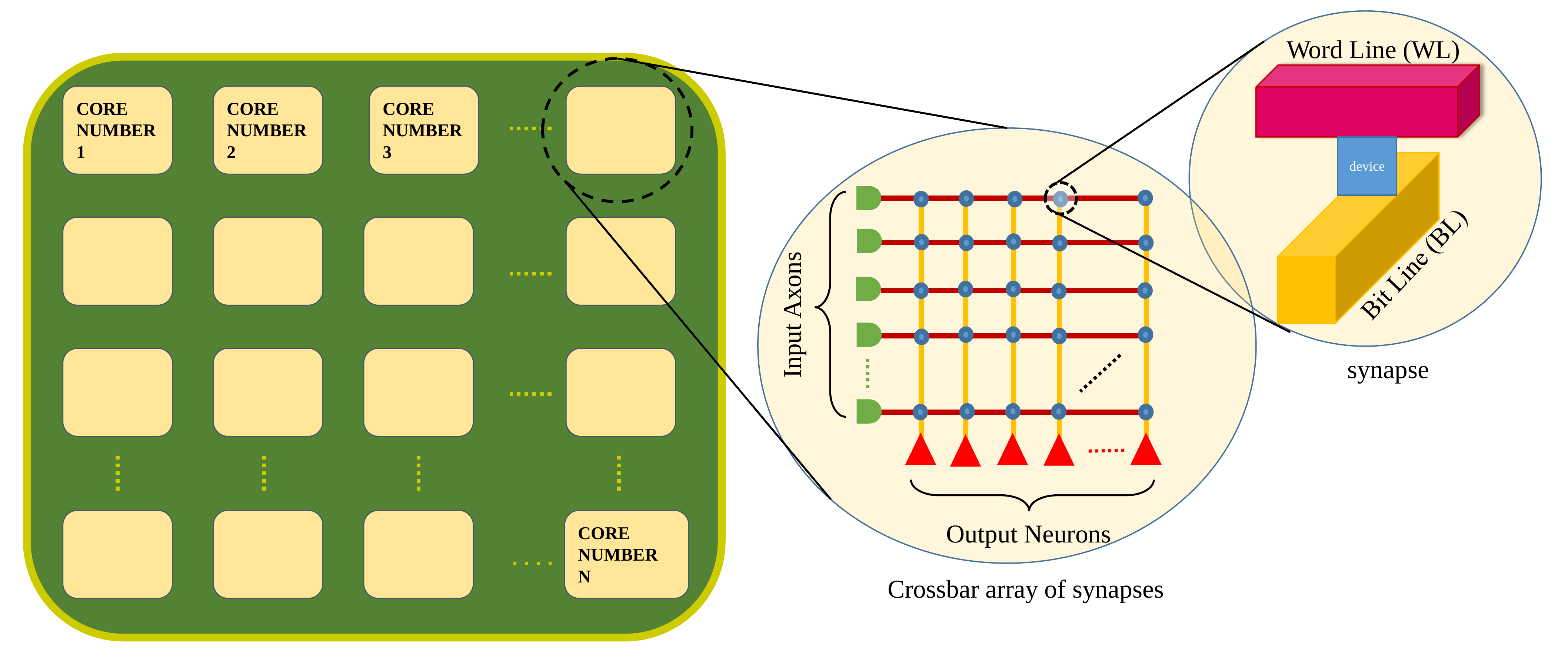}% This is a *.eps file
\end{center}
\caption{ A schematic of a neuromorphic chip with N number of neuromorphic cores. First inset shows the crossbar array of synapses within each core. A memory device is used to implement each synapse at the crossbar intersection (as shown in second inset).}
\label{fig:1}
\end{figure}

A schematic of a neuromorphic chip is shown in fig \ref{fig:1}. The chip has N number of neuromorphic cores. Network on chip (NoC) or router interfaces are not shown for illustration purposes. Each neuromorphic core contains a crossbar array of synapses as shown in the first inset of the figure. The rows and columns of the crossbar correspond to input axons and output neurons respectively. These axons and neurons are interconnected to each other at their intersection. Within each intersection of the crossbar between the word line and the bit line, is a synaptic device which has memory and can perform some in-memory computation (as shown in the second inset). The crossbar architecture is further discussed in subsection \ref{Crossbar}. Considering such a neuromorphic chip, there are several hardware constraints, namely, low bit precision of synaptic weights and output activations \cite{NC_constraints}, synaptic noise and variability \cite{RRAM_variability, RRAM_RTN}, number of neuromorphic cores, the size of each core size in a neuromorphic core and fan in, fan out degree of each neuron \cite{NC_constraints, MAD_Arxiv}. 

The aim of this paper is as follows:
\begin{itemize}
\item to design a hardware friendly deep neural network architecture in order to fit onto a neuromorphic hardware with limited number of cores;
\item to maximize the utilization of a single neuromorphic core with limited core size;
\item to study how the selected architecture would work with different core sizes, and its corresponding classification accuracy. 
\end{itemize}
Towards this aim, we propose a novel neural network architecture based on existing convolution techniques. We consider a particular architecture to be hardware-friendly so long as it is able to be mapped onto a neuromorphic chip, while achieving a reasonable level of classification accuracy. In this regard, one may take two approaches, either design the network from scratch within the constraints of the neuromorphic hardware specifications or optimise an existing deep network taking into account the hardware specifications. Optimizing an existing deep network can be done by reducing the number of features in each layer so as to fit onto the neuromorphic core without having to split the convolution matrix among different cores. The novel neural network proposed is obtained by extracting different layers from different existing architectures and further modifying the features of some of the layers in these architectures so as to fit onto the neuromorphic hardware. 

The paper is organised as follows. Section \ref{M&M} describes the types of convolutions present in a CNN, a short review of different CNN architectures, MaD framework and computation in a crossbar array. In the subsequent section \ref{PA}, the proposed architecture is illustrated. Section \ref{result} talks about the results obtained and the paper then concludes with section \ref{Conc}.

%Hardware resource utilization is calculated using the Python wrapper proposed in our previous work \cite{MAD_Arxiv}. 

\section{Materials and Methods}
\label{M&M}

%For requirements for a specific article type please refer to the Article Types on any Frontiers journal page. Please also refer to  \href{http://home.frontiersin.org/about/author-guidelines#Sections}{Author Guidelines} for further information on how to organize your manuscript in the required sections or their equivalents for your field

% For Original Research articles, please note that the Material and Methods section can be placed in any of the following ways: before Results, before Discussion or after Discussion.

%\section{Manuscript Formatting}

\subsection{Types of convolution used in a convolutional neural network(CNN)}
\label{convs}

Convolution is an operation that involves the summation of product of terms in two matrices. Convolution and its application in a neural network is to extract the features of the input across each layer of the network. The different convolution techniques applied in a CNN are discussed below. 

\subsubsection{Standard convolution}

\begin{figure}[h!]
\begin{center}
\includegraphics[width=16cm]{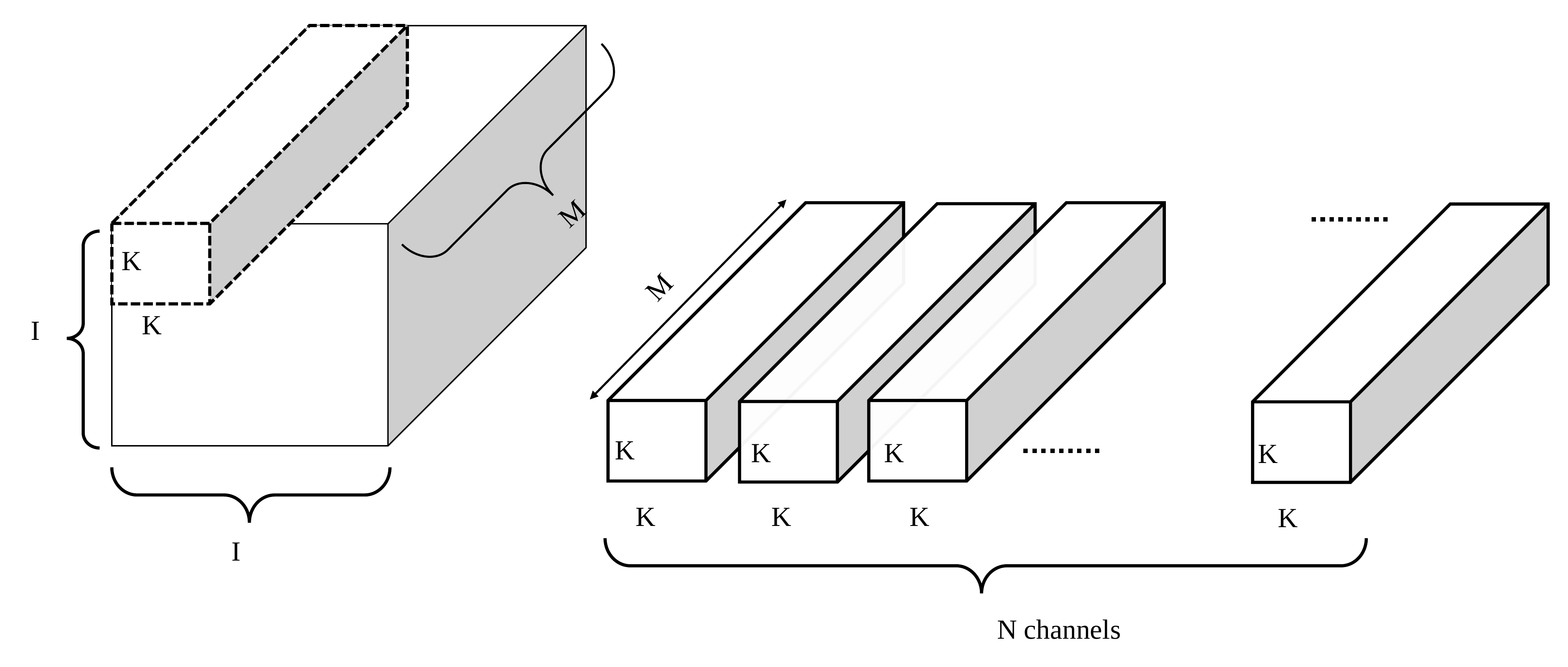}
\end{center}
\caption{Illustration of the standard convolution.}
\label{fig:std_conv}
\end{figure}

In the standard convolution, each filter/kernel is multiplied and summed across the whole feature map (input channels). Consider the convolution operation on an input matrix of I $\times$ I $\times$ M with a filter of K $\times$ K $\times$ M (as shown by the dotted lines in fig. \ref{fig:std_conv}). This filter operation will give one output feature map which shares filter values for each stride of the convolution. In order to generate N such feature maps, different filters are used, as shown in fig. \ref{fig:std_conv}. The computational complexity C of a normal convolution is given as below:
\begin{equation}
C = O^2 * K^2 * M * N
\end{equation}
where, O = output size after the convolution \\
K = filter size \\
M = number of input channels \\
N = number of output channels

The standard convolution can be computationally intensive and also hard to map onto neuromorphic cores of limited sizes. Therefore, computationally less intensive convolution techniques have been proposed, which include the pointwise convolution, depthwise convolution, flattened convolution, group convolution etc., some of which are further discussed below.

\subsubsection{Pointwise convolution}

The pointwise convolution is a subset of the standard convolution whereby the filter size per channel is set to 1 $\times$ 1. The entire filter size is hence 1 $\times$ 1 $\times$ M $\times$ N, where M is the number of input channels and N, the number of output channels. Since the filter size is reduced, the computational complexity is also reduced by an order of the square of the filter size. Its computational complexity, C is given as below:
\begin{equation}
C = O^2 * M * N
\end{equation}
where, O = output size after convolution \\
M = number of input channels \\
N = number of output channels  

\subsubsection{Depthwise convolution}

In depthwise convolution, the convolution is independently applied to each input channel so as to obtain its corresponding output feature map. Hence, in depthwise convolution, the summation operation is applied within each corresponding input channel. After the depthwise convolution operation, the number of output channels obtained is the same as that of the number of input channels. One may however also increase the number of output channels per input channel using a depth multiplier. 

\begin{figure}[h!]
\begin{center}
\includegraphics[width=17cm]{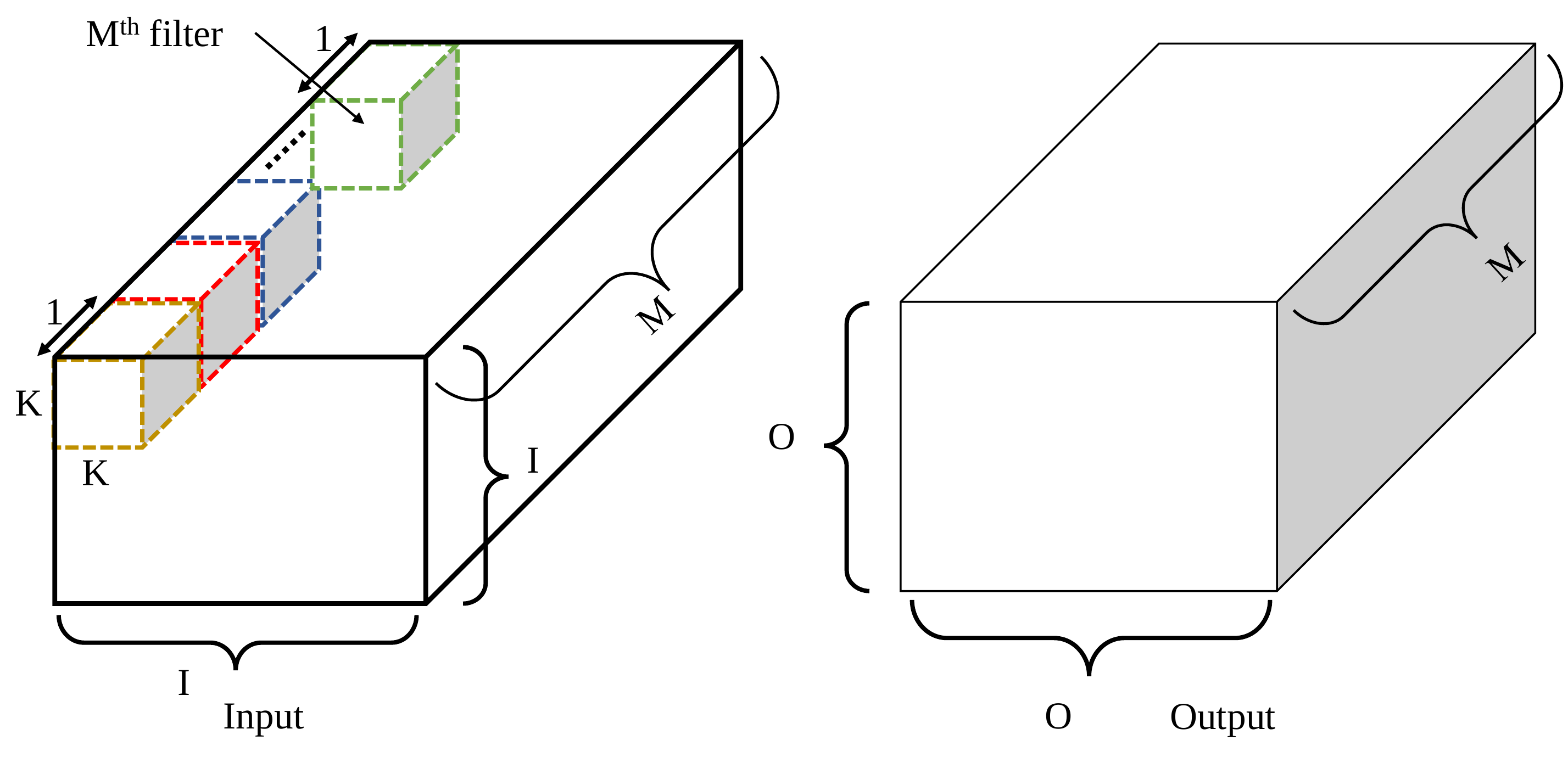}
\end{center}
\caption{Illustration of the depthwise convolution. Note that the depth multiplier is set to one here.}
\label{fig:dp_conv}
\end{figure}

As shown in fig. \ref{fig:dp_conv}, the input matrix is convolved with M different filters, each of size K $\times$ K. The output of each depthwise convolution involving a filter and a single input channel is O $\times$ O $\times$ 1, and M such filters compute an output of dimensions O $\times$ O $\times$ M. The depth multiplier is set to one here. The computational complexity, C of the depthwise convolution considering depth multiplier is given as below:
\begin{equation}
C = O^2 * K^2 * M * D
\end{equation}
where, O = output size after convolution \\
K = filter size \\
M = number of input channels \\
D = depth multiplier

The application of the depthwise convolution, then pointwise convolution together, is known as the depthwise separable convolution. Depthwise separable convolution is used extensively in MobileNets \cite{MobileNet}. 

\subsubsection{Grouped convolution}

\begin{figure}[h!]
\begin{center}
\includegraphics[width=17cm]{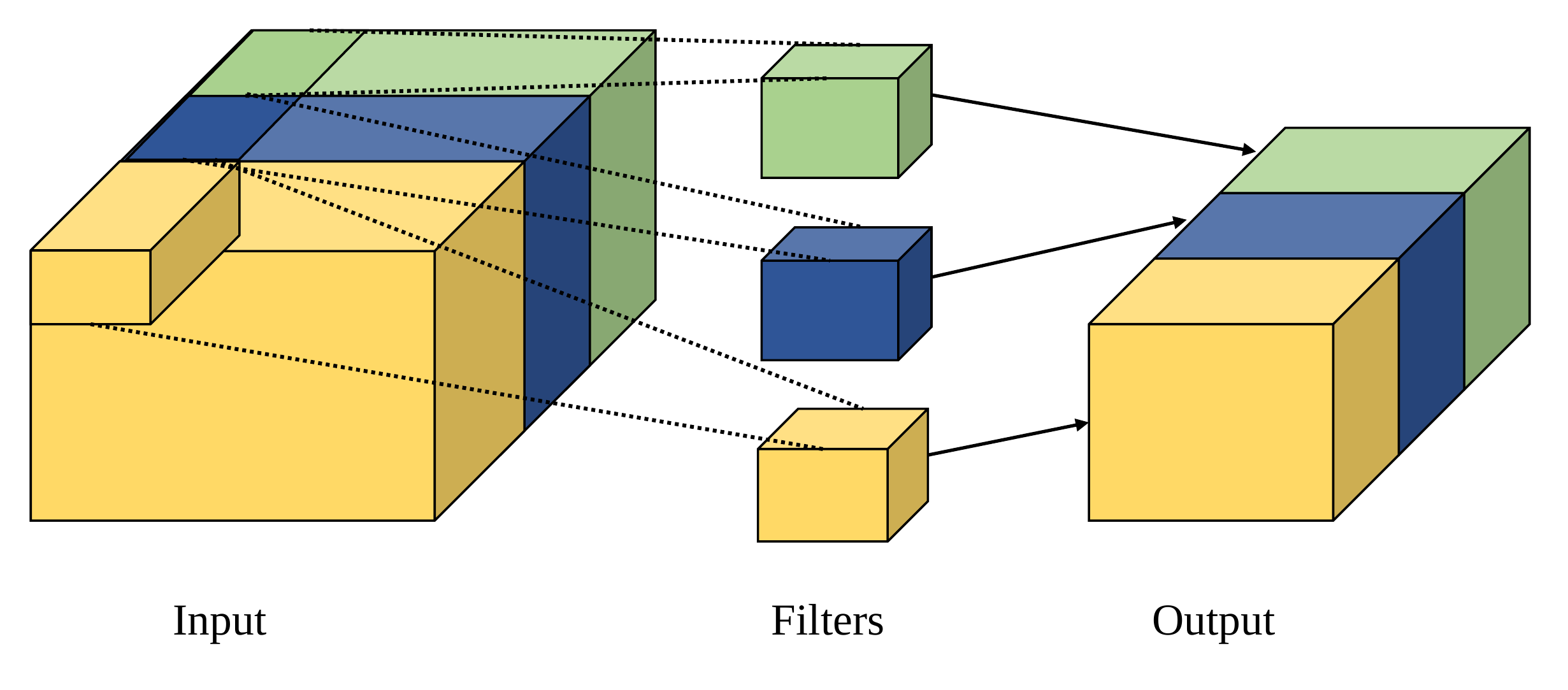}
\end{center}
\caption{Illustration of grouped convolution.}
\label{fig:group_conv}
\end{figure}

Grouped convolution is another convolution technique whereby the standard convolution is applied separately to an input matrix diced into equal parts along its channel axis. As shown in fig. \ref{fig:group_conv}, the input channels and filters are divided into several equal parts along the channel axis. All the separate output features of these independent parts are then combined into a final output. Depending on how the parts are combined, different variations of the grouped convolutions include: stacked convolution, dependent stacked convolution and shuffled convolution. Computational complexity of the grouped convolution is calculated as per the standard convolution but grouped convolutions are more neuromorphic hardware friendly as each neuron has a lower fan in/fan out degree when mapped onto a crossbar array of synapses.

%\subsubsection{Flattened convolution}

%There are 5 heading levels

\subsection{Different convolutional neural networks}
\label{Evo}

This subsection briefly introduces several popular CNNs. These are some of the CNNs we have studied whilst proposing the novel CNN architecture that is more hardware-friendly. 
%(We consider only CNNs that do not have connections that skip layers such as the residual networks. Feedforward neural network architecture is much more appropriate for its implementation on a neuromorphic chip with crossbar array of synapses. cys: i recommend removing these lines in this bracket, as residual networks can be implemented in a chip, just that it will increase fan-in/fan-out degree. i have mentioned this as future work in the conclusion)

\begin{figure}[h!]
\begin{center}
\includegraphics[width=17cm]{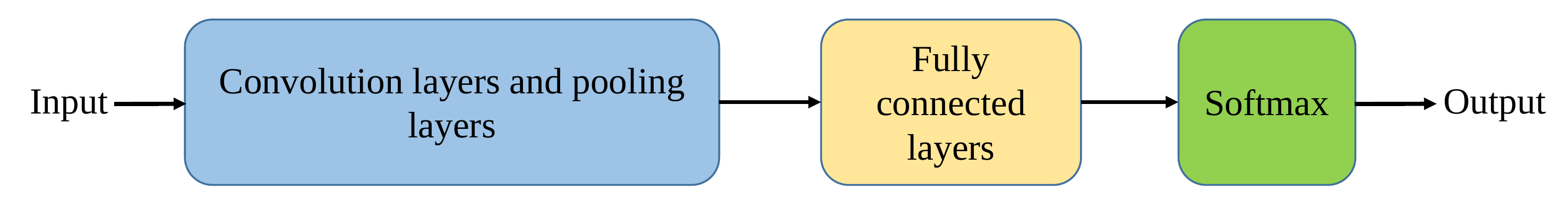}
\end{center}
\caption{Block diagram illustration of the AlexNet and VGGNet.}
\label{fig:VGG}
\end{figure}

\subsubsection{AlexNet}

Geoffrey Hinton and team introduced the first deep neural network architecture, Alexnet, which was named after his student. AlexNet \cite{AlexNet} paved the way for a new field of artificial intelligence research known as deep learning. Deep learning has since been applied to many fields in AI, with state-of-the-art results \cite{Fractional_max, ELU, DropConnect}. Figure \ref{fig:VGG} details the architecture of the AlexNet in a block diagram. AlexNet is a CNN with alternate convolutional and pooling layers, ending with fully connected layers.

\subsubsection{VGGNet}

VGGNet \cite{VGGNet} was introduced by the Visual Graphics Group at Oxford. VGGNet is similar to AlexNet, with slight modification in the placement of layers. The architecture is deeper compared to AlexNet, and retains the input width in the early layers. Figure \ref{fig:VGG} is also representative of the VGGNet architecture. We note that in VGGNet, the pooling layers are applied after every two convolution layers in the beginning of the architecture and every three layers afterwards. This technique preserves the larger width of the layers at the beginning of the architecture. VGGNet uses three layers of full connections before the softmax classifier.

\subsubsection{GoogLeNet}

\begin{figure}[h!]
\begin{center}
\includegraphics[width=17cm]{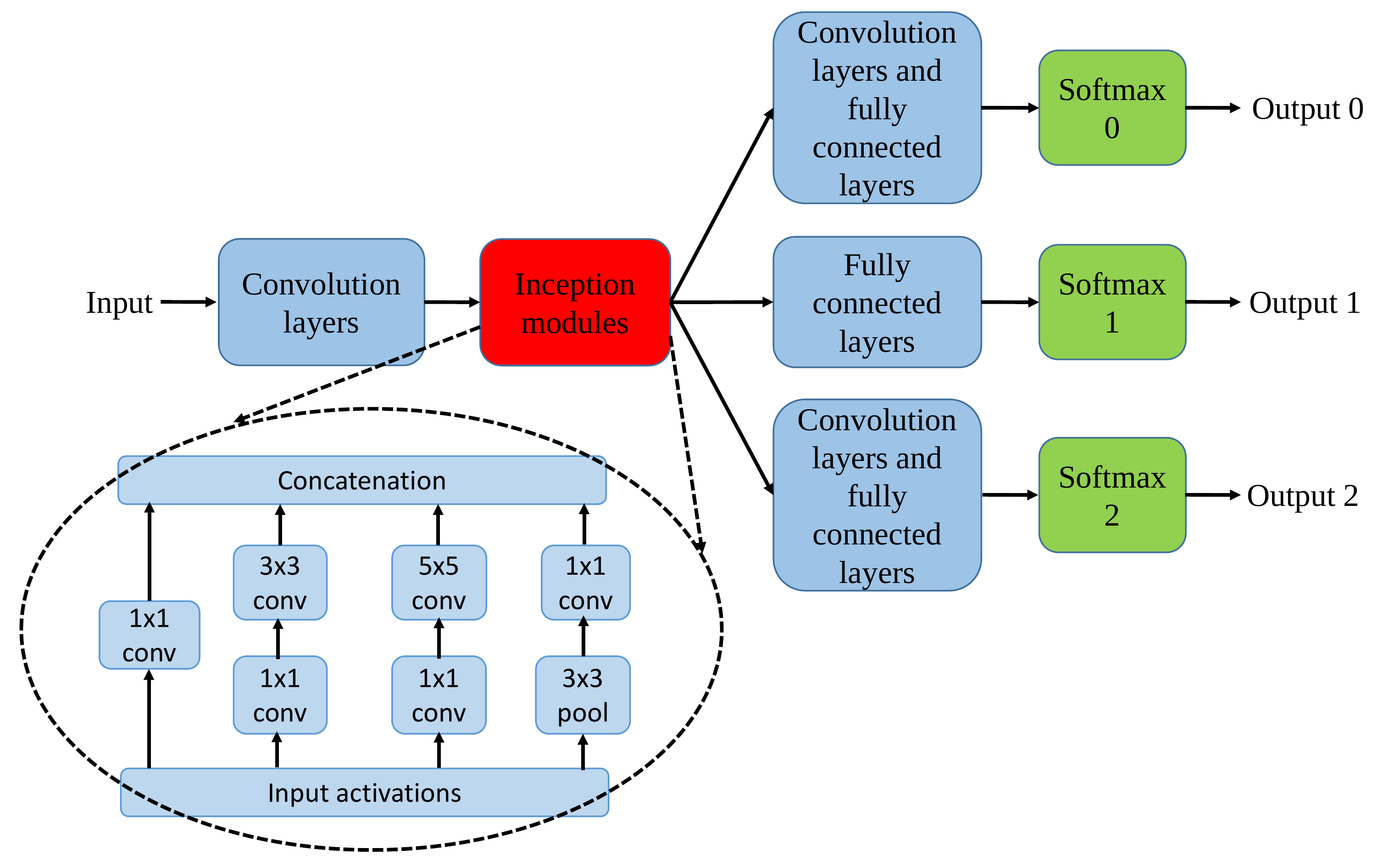}
\end{center}
\caption{Block diagram illustration of GoogLeNet.}
\label{fig:GoogLeNet}
\end{figure}

GoogLeNet \cite{GoogLeNet} is a 22 layer deep CNN. Its main novelty is the inception module, which concatenates the output of several convolutions of the same activations from the previous layer. Figure \ref{fig:GoogLeNet} shows the blockwise architecture of the GoogLeNet and the inception module in the inset. GoogLeNet started using 1$\times$1 convolution or pointwise convolution extensively in their inception modules, which was introduced in \cite{NIN}. It is the winner of the ILSVRC 2014 classification challenge which involves classifying the IMAGENET dataset \cite{imagenet}.

\subsubsection{SqueezeNet}

\begin{figure}[h!]
\begin{center}
\includegraphics[width=17cm]{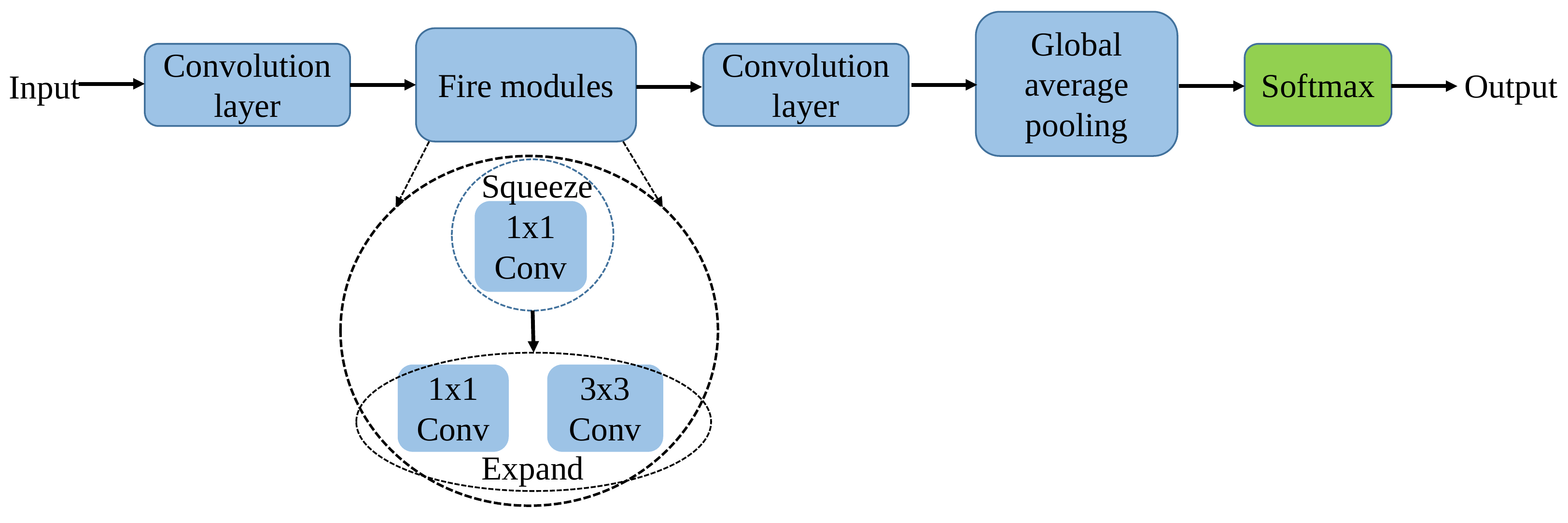}
\end{center}
\caption{Block diagram illustration of SqueezeNet.}
\label{fig:SqueezeNet}
\end{figure}

SqueezeNet \cite{SqueezeNet} is another architecture similar to MobileNet with an even smaller size. One of the compressed versions of SqueezeNet has only 0.47MB of parameters, making it ideal for deployment on a mobile platform. SqueezeNet has a fire module, consisting of a squeezing network module followed by an expanding network module. Fire module is the key to preserving classification accuracy while reducing network size. Figure \ref{fig:SqueezeNet} shows the details of the SqueezeNet architecture. It does not have any fully connected layers before the softmax classifier, in contrast to AlexNet and VGGNet.

\subsubsection{MobileNet}

\begin{figure}[h!]
\begin{center}
\includegraphics[width=17cm]{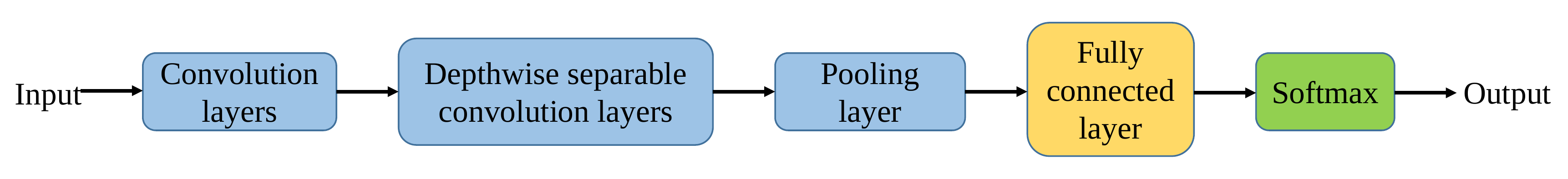}
\end{center}
\caption{Block diagram illustration of MobileNet.}
\label{fig:MobileNet}
\end{figure}

MobileNet \cite{MobileNet} was proposed by Google, with the main motivation of reducing the the network size and its parameters. MobileNet uses depthwise separable convolution as illustrated in the previous section \ref{convs}. Figure \ref{fig:MobileNet} shows the architecture of the MobileNet. It is also more neuromorphic hardware-friendly as the fan in/fan out connections of each neuron is much less, as a result of the pointwise and depthwise convolutions.

%\subsubsection{EffNet}
%(cys: you do not want to discuss effnet?)
\subsection{Crossbar array of synapses}
\label{Crossbar}

The neuromorphic chip discussed in this paper is based on a crossbar architecture \cite{crossbar} of non-volatile memory synapses. In Fig. \ref{fig:1}, the first inset shows a crossbar architecture of synapses in a neuromorphic chip. For a biological neuron, an axon connects the pre-synaptic neuron to the synapse, which is the site of connection between the axon of the pre-synaptic neuron and the dendrite of the post-synaptic neuron. Similarly, in CNNs on a neuromorphic hardware, the synapse can be viewed as the site of connections between the input neurons and output neurons of a convolution layer. Memory device is used to represent these synaptic weights which are analogous to the weights in the filters of the CNNs (Fig. \ref{fig:1}, second inset). In the mesh-like crossbar array, the synapse of the neuromorphic core establish connections between axons and neurons of that neuromorphic core. Typically in a neuromorphic chip, spiking neurons are used to integrate the current from the synapses and a spike is emitted, when the firing threshold is met. Hence, each neuron at the bottom of the crossbar array performs a nonlinear function on the convolution operation between input and synaptic weights. These operations are also termed as matrix dot vector multiplications (MVM) \cite{MVM}.

\subsection{Python wrapper: MaD Framework}
\label{EF}

One of the main challenges in the field of neuromorphic hardware is to efficiently map the neurons in a CNN onto the neuromorphic chip while fulfilling hardware constraints such as core size, number of cores and fan-in/fan-out \cite{neutrams}. Existing neuromorphic chips have a mapping framework which is hardware specific. IBM's TrueNorth chip \cite{truenorth_TCAD} uses Corelet language \cite{IBM_corelet} based on MATLAB, a programming language specific to their hardware. Within this MATLAB framework, a mapping technique is integrated as a minimization problem \cite{truenorth_TCAD}. SpiNNaker and BrainScaleS uses a simulator-independent language, PyNN \cite{PyNN}, based on Python. Sequential mapping is used in SpiNNaker. Neural engineering framework (NEF) is developed for Neurogrid \cite{NEF}. Neutrams \cite{neutrams} addresses an optimized mapping technique based on graph partition problem: Kernighan-Lin (KL) partitioning strategy for network on chip (NoC). Even though, every neuromorphic hardware simulator tool provides certain mapping techniques, optimized mapping onto a single neuromorphic core is often neglected and is relatively unexplored. While developing deep neural networks that are to be mapped onto a neuromorphic chip, one need not in principle be aware of the underlying hardware constraints of the chip. However, to better utilize the chip for a classification task, software and hardware co-design is encouraged, which requires the neural network designer to be aware of the underlying hardware constraints. The aforementioned issues are addressed using the MaD framework. The MaD framework is a generic Python wrapper which has an optimized algorithm for mapping a feed-forward neural network such as the MLP, CNN and spiking neural network (SNN) onto a crossbar array of synapses with corresponding synaptic weights, thereby fitting the neurons using the least number of neuromorphic cores. The Python wrapper is also suitable as a debugging tool for verification of inferencing results after mapping the neural network architectures onto the neuromorphic chip. Thus, together the framework is called the mapping and debugging (MaD) framework. This Python wrapper is developed in connection with the simulator in \cite{neurosim}, which shares several similar techniques to that of Neutrams \cite{neutrams}.

\begin{figure}[htbp]
\centerline{\includegraphics[width=11cm,height=7cm]{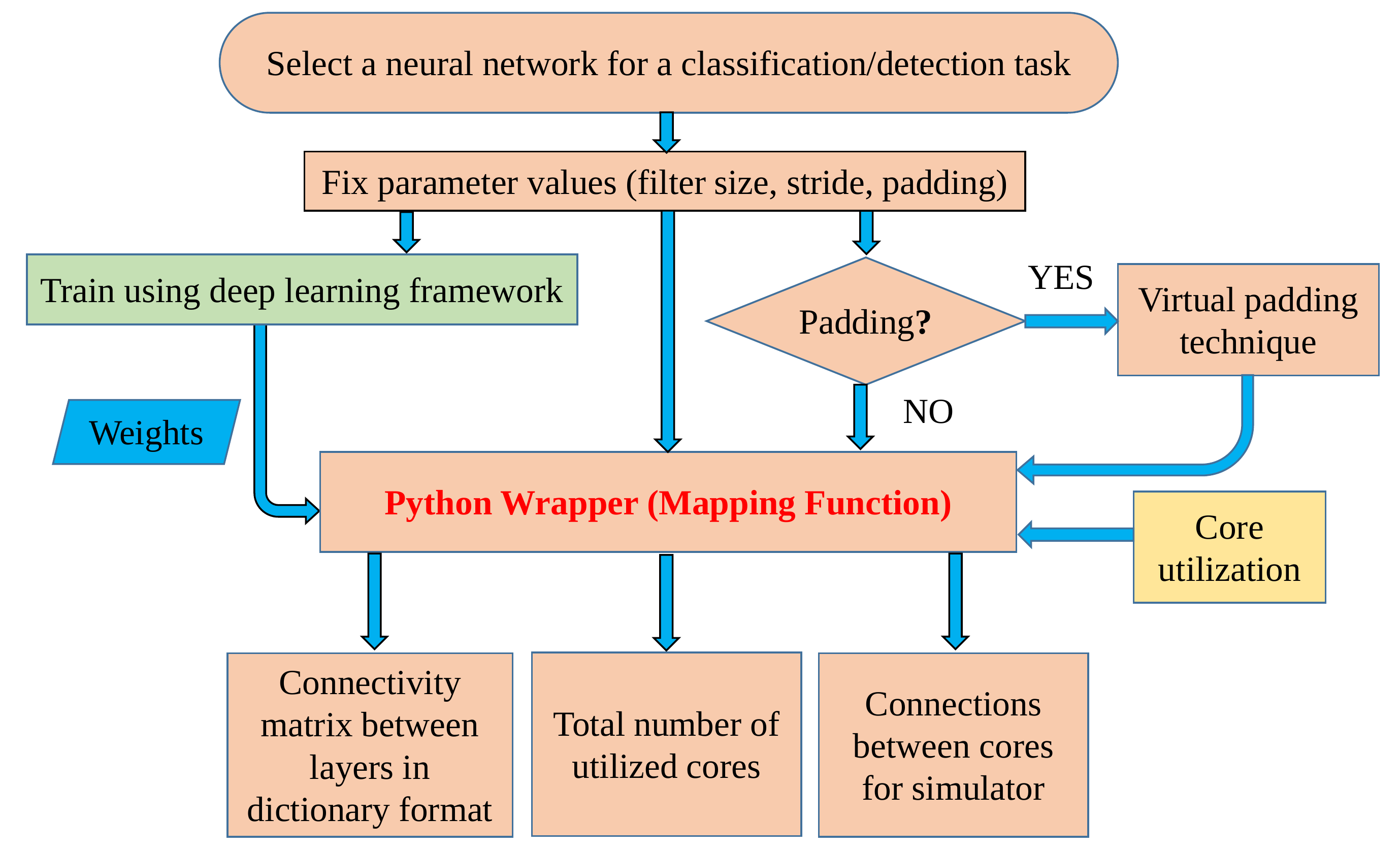}}
\caption{\textbf{Flowchart of Python wrapper:} Details of the Python wrapper shown in a flowchart, with the input and output of the mapping function shown. Core utilization and weight files are in different color to denote that these inputs are results from other functions.}
\label{Flowchart}
\end{figure}

\begin{figure}[htbp]
\centerline{\includegraphics[scale=0.5]{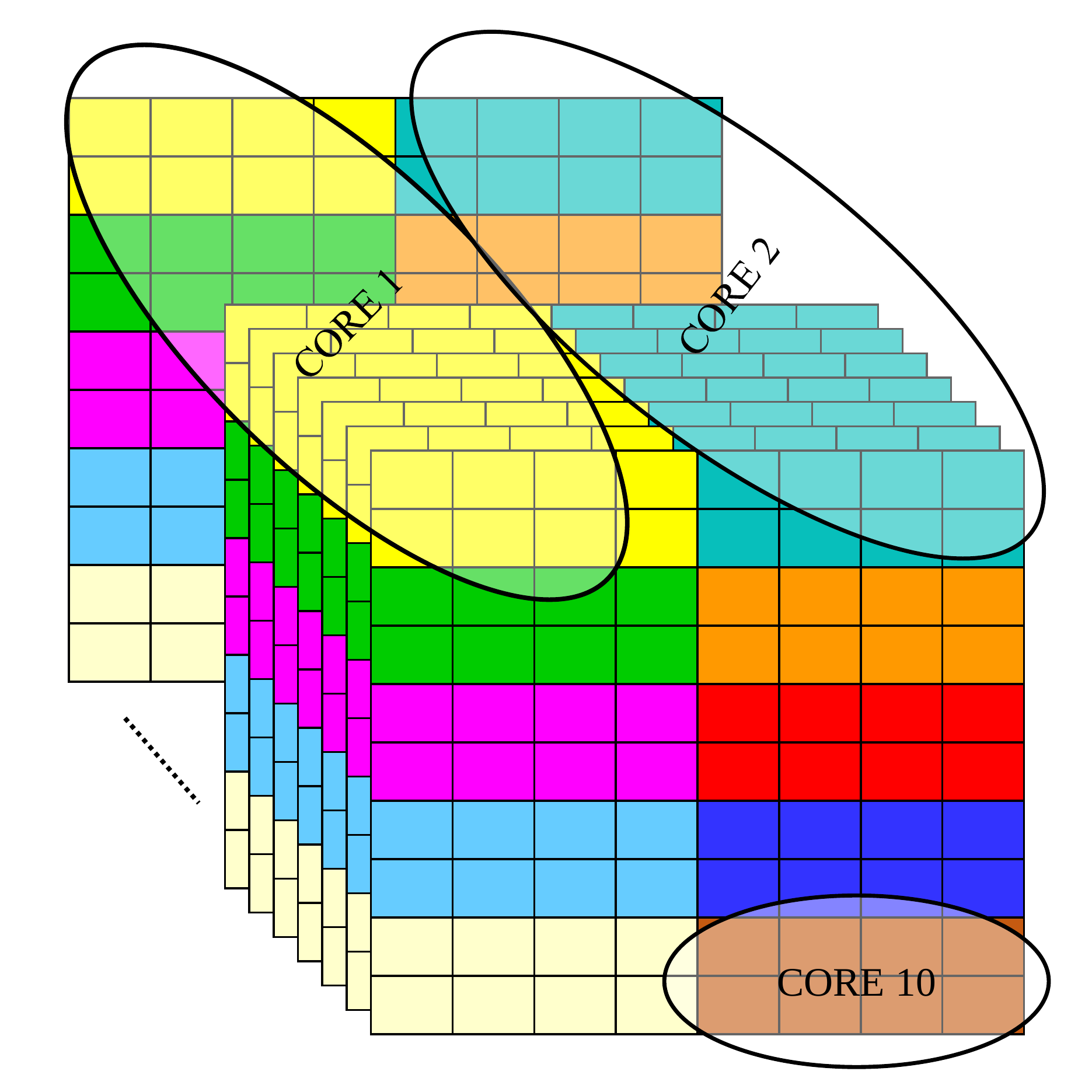}}
\caption{Division of a CNN layer into different neuromorphic cores.}
\label{Cores_layer}
\end{figure}

The functionalities of the MaD framework is explained in the flowchart (fig. \ref{Flowchart}). Given a CNN chosen for a classification or detection task, its hyper-parameters such as filter size, strides and padding at each layer is known. The chosen network is trained using existing deep learning frameworks  and the trained weight variables (together with the above-mentioned hyper-parameters) are input into the mapping function. Core utilization is defined as the number of axons and neurons utilized in a single neuromorphic core, represented as [axons $\times$ neurons]. Core utilization, as shown in the flowchart, is output from another function which calculates the number of axons and the number of neurons used for mapping a section of a particular layer onto a single core. The mapping of different portions of a convolutional layer onto different cores is shown in the fig. \ref{Cores_layer}. Different colors within the layer show that these neurons are mapped onto different cores. For example, neurons in yellow are mapped onto core 1 and neurons in brown are mapped onto core 10 etc. The details of the mapping function, core utilization and padding techniques are given in subsequent subsections. The supplementary material contains the discussion on the optimizations performed during mapping.%This section concludes by discussing the optimizations performed during mapping.

%\subsection{Flowchart}
%steps for creating MaD 

\subsubsection{Mapping Function}
\label{MF}

The mapping function is the core of the Python wrapper. Fig. \ref{Flowchart} also shows the input and output of the mapping function. The inputs to mapping function are input size, filter size, stride, padding, core utilization and weight files. The input size is the size of the input datasets, for eg. 28$\times$28 in the case of MNIST or 32$\times$32 in the case of CIFAR-10. Filter size is the size of filters used for each convolution layer. For instance, it is 3$\times$3 throughout all convolutional layers in the CNNs described in result subsection given in supplementary material. Stride and padding are layer-specific. The detailed calculation of the core utilization is discussed in subsection \ref{CU}. Weight files are the weights obtained after training the deep network. The output section in fig. \ref{Flowchart} shows the outputs of the mapping function. There are mainly three outputs, a connectivity matrix for verifying the interconnectivity between the cores and within the core, to verify the cores utilized and an automated generation of connection list for simulator.

The steps for mapping are as follows:
\begin{itemize}
\item Name the neurons using the convention: L1-F1-N[1,1], which implies layer:1, feature map:1, and neuron in row:1 and column:1.
\item Create a connectivity list stating how populations of neurons in one layer are connected to populations in the previous layer.
\item Choose a population of neurons from a particular layer, based on the core utilization, to be mapped on to a particular core.
\item Repeat step 3 until all neurons are mapped onto a core. Since the naming and connectivity list are determined at the beginning, the neurons and axons are automatically duplicated across cores during mapping.  
\end{itemize}
     
\subsubsection{Core Utilization}
\label{CU}

\begin{figure}[htbp]
\centerline{\includegraphics[width=11cm,height=7cm]{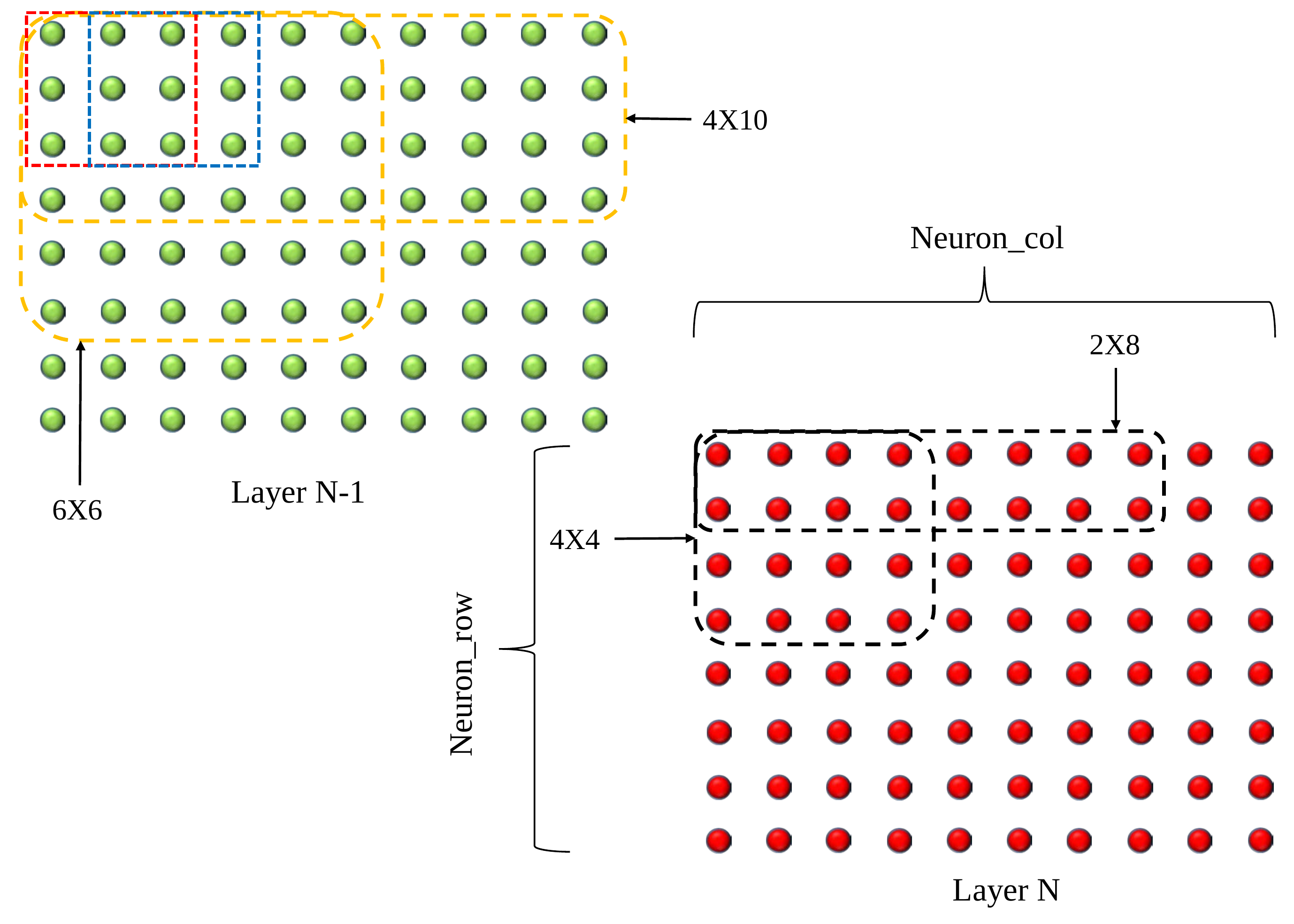}}
\caption{Two layers of the convolution operation to illustrate optimization of core utilization. Layer N-1 neurons are in green, while layer N neurons are in red. Synaptic connections are shown for two neurons in layer N.}
\label{Core_uti}
\end{figure}
Consider two layers of a CNN as shown in fig. \ref{Core_uti}. First two neurons in layer N (in red) are connected to layer N-1 neurons (in green) with the synaptic connections as shown with their respective afferent neurons in red and blue squares. Size of the convolution filter used is 3$\times$3. The synaptic connections extend across the layers as per the kernel size and strides used during convolution.  While mapping these two layers in fig. \ref{Core_uti} onto a core with crossbar array, the green neurons in layer N-1 will be the axons and the red neurons in layer N will be the neurons as in fig. \ref{fig:1} (notice the overlap of filter kernel as it traverses across the layer N-1). These overlapping green neurons may however be mapped onto the axons of the crossbar array without duplicates. Duplication of axons is not desirable. Duplication of axons while mapping onto a neuromorphic core will require an input to be duplicated into multiple axons within a neuromorphic core (increases core usage). Hence, the toeplitz matrix method is utilized for efficient mapping of these convolution layers onto a neuromorphic core \cite{toeplitz_IBM} \cite{toeplitz}. For a given mapping on a particular core, the core utilization maybe calculated based on the number of neurons and axons connected together. The number of axons can be evaluated as an algorithmic condition in the mapping function as there are overlapping axons whereas neurons selection become bit straight forward. The overlapping axons are defined as the axons (in layer N-1) which share connections with more than a single neuron (in layer N), the term overlapping is because of the overlapping nature of the axons with the neighbourhood of the kernel filter with respect to strides (see layer N-1 in fig. \ref{Core_uti}, the overlapping axons from the red and blue dotted squares are 6). Depending on this overlap, kernel filter size and strides, the total number of axons to be selected is given by the below equation:

\begin{equation}
\begin{aligned}
\label{eq:Axons}
N\_axons = K*K + K*S(Neuron\_col-1) + \\
		   S*S(Neuron\_col-1)(Neuron\_row-1) + \\
		   K*S(Neuron\_row-1)
\end{aligned}
\end{equation}
Where, \\
$N\_axons$ = total number of axons to be selected \\
K = convolution kernel filter size \\
S = stride \\
$Neuron\_row$ = number of neurons across row \\
$Neuron\_col$ = number of neurons across column \\
The selection of neurons, $Neuron\_row$ and $Neuron\_col$, in a layer has to satisfy the condition: number of axons, $N\_axons$ $<=$ number of physical axons (eg. 256 or 512 or 1024) in the neuromorphic core. Eq. \ref{eq:Axons} considers only a single feature map; this can be easily extended to multiple feature maps by multiplying with the respective number of feature maps.

\subsection{Computation in a crossbar array}

\begin{figure}[h!]
\begin{center}
\includegraphics[width=17cm]{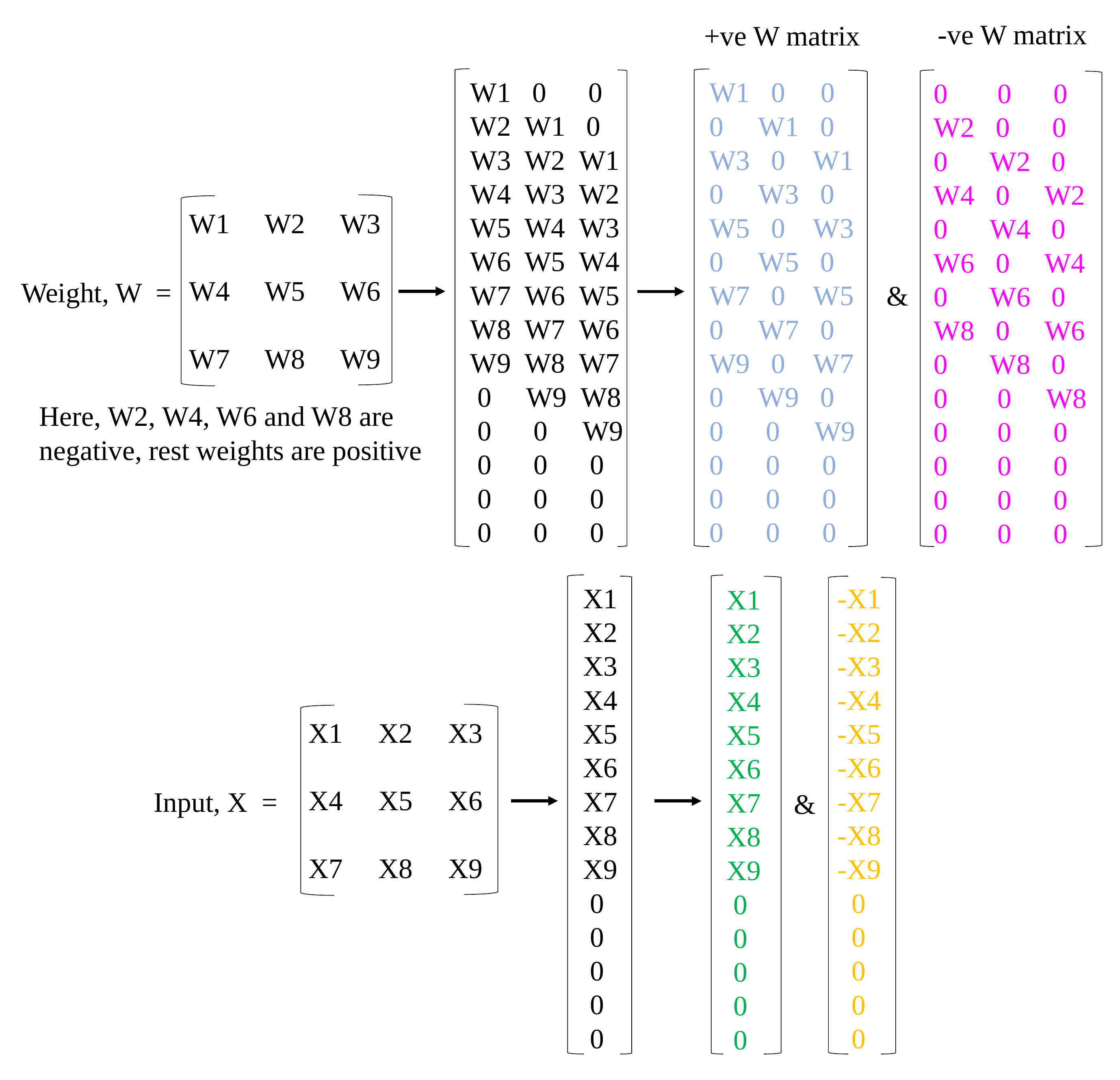}
\end{center}
\caption{Division of network parameters, weights and input activations into positive and negative matrices.}
\label{fig:Crossbar_weights}
\end{figure}

\begin{figure}[h!]
\begin{center}
\includegraphics[width=17cm]{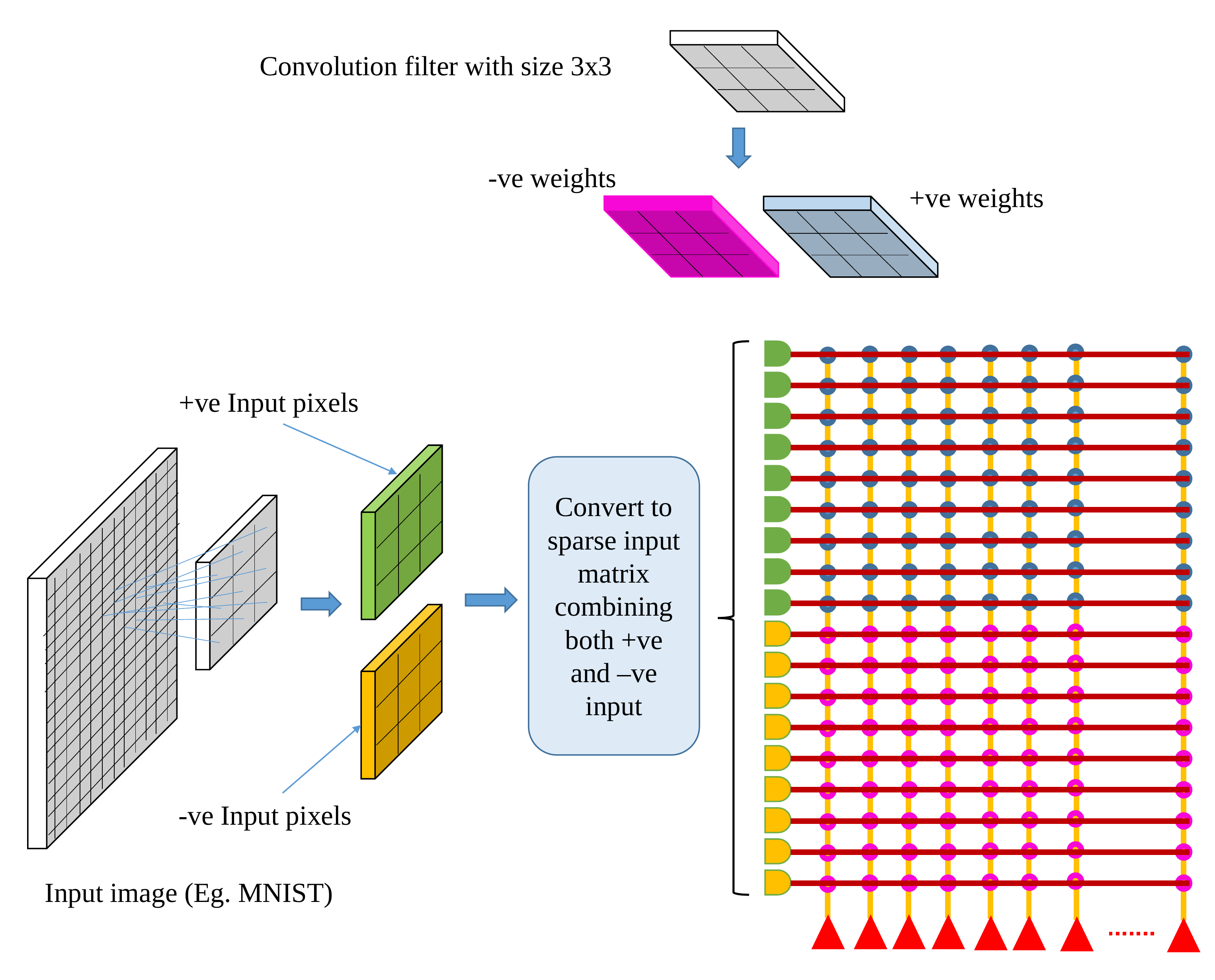}
\end{center}
\caption{Computation in a crossbar architecture within a single neuromorphic core.}
\label{fig:Crossbar_com}
\end{figure}

The crossbar array of synapses in a neuromorphic chip is capable of doing convolutions. Mathematically, convolution is the sum of dot product of two input matrices. One matrix being the input matrix and the another is the filter matrix as shown in fig. \ref{fig:Crossbar_weights}. In CNNs, the input matrix will be the activations from the prior layer while the filter matrix is the convolution filter kernel, saved as weights, W after a CNN is trained. Since these weights can be either positive or negative values after training, one way of implementing convolution on a crossbar array is to split the weights into positive and negative matrices along with two copies of input matrices in positive and negative values. The details of the matrix generation is shown in fig. \ref{fig:Crossbar_weights}, which incorporate the convolution operation in crossbar arrays as described in \cite{Alom_2016} (also referred in our previous paper \cite{Roshan_RRAM_ARXIV}). Single column of crossbar gives the output of a convolution operation, which is the output of corresponding neuron. Convolution operation is extended to multiple columns of RRAM synapses to compute in parallel. This requires the weights and inputs to be represented in a toeplitz matrix, as shown in fig. \ref{fig:Crossbar_weights}. \cite{Alom_2017} illustrated such an implementation in fig. \ref{fig:Crossbar_com}. This implementation doubles the utilization of hardware resources which is similar to IBM Truenorth \cite{IBM_Esser}, where they need two synapses to implement the ternary weights (-1, 0, +1). 

In order to mitigate the above described double synaptic utilization in a neuromorphic hardware, one can implement two memory devices at each synapse to represent both positive and negative weights by subtraction. %Hence, the mapping of a deep neural network onto the neuromorphic hardware with crossbar array of synapses will be straightforward and was discussed in our earlier paper \cite{MAD_Arxiv}.
This implementation does not need to partition the weights and inputs into positive and negative matrices instead, generating a toeplitz matrix is sufficient.

\section{Proposed architecture}
\label{PA}

\begin{figure}[h!]
\begin{center}
\includegraphics[width=17cm]{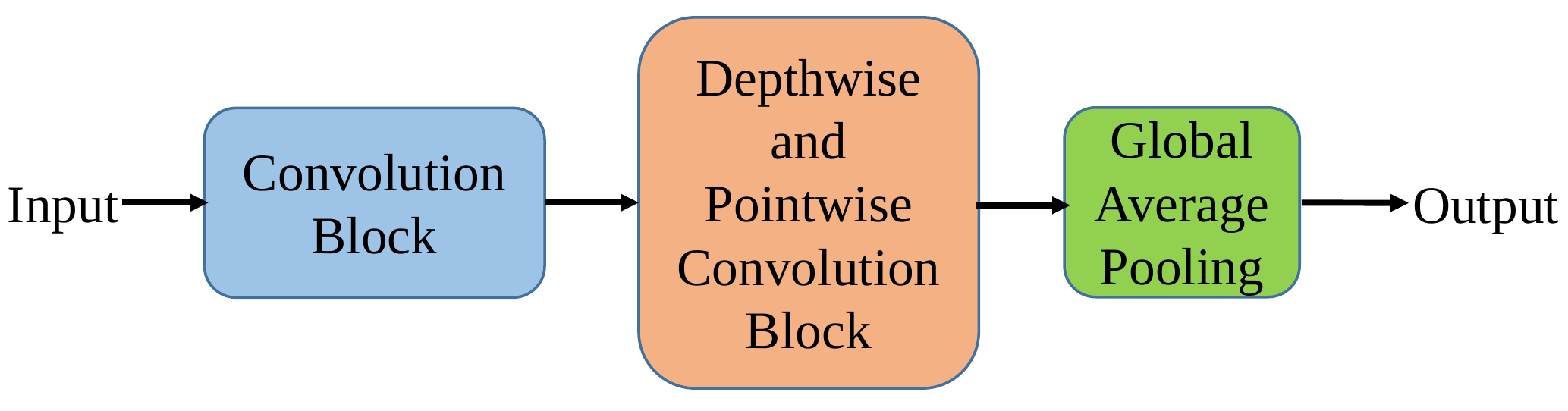}% This is a *.eps file
\end{center}
\caption{ The proposed hardware-friendly neural network architecture in block diagram.}
\label{fig:Arch}
\end{figure}

The proposed architecture borrows from the different CNNs discussed in section \ref{Evo}. It is a hybrid combination of the VGGNet, MobileNet and SqueezeNet. As shown in fig. \ref{fig:Arch}, the first three layers are convolutional layers as in the case of VGGNet (convolutional block) and the next layers are alternate layers of Depthwise and Pointwise convolutions (Depthwise and Pointwise convolutional block) as in the case of MobileNet. Since the fully connected layers require more parameters and have large fan in/fan out degrees, the last fully connected layer of MobileNet is replaced with global average pooling, similar to the SqueezeNet architecture. Pooling layers are not necessary in a CNN. It can be replaced by using convolutional layers with stride of 2 so as to achieve dimension reduction without significant loss in accuracy, even though mathematically, they are different \cite{without_pool}. Thus, the proposed architecture is novel whereby it does not have pooling and fully connected layers. The detailed input size and output size of each layer of the proposed architecture for different core sizes are given in table \ref{tab1}.

\begin{table}[htbp]
\caption{Neural network architecture (NN archi.) for different core sizes}
\begin{center}
\begin{tabular}{|c|c|c|c|c|c|c|}
\hline
\textbf{NN}&\multicolumn{6}{|c|}{\textbf{Core Size}} \\
\cline{2-7} 
 \textbf{archi.} & \multicolumn{2}{|c|}{\textbf{\textit{256$\times$256}}} &\multicolumn{2}{|c|}{\textbf{\textit{512$\times$512}}} & \multicolumn{2}{|c|}{\textbf{\textit{1024$\times$1024}}} \\
\cline{2-7} 
 \textbf{Layers} & \textbf{Input Size} & \textbf{Output Size} & \textbf{Input Size} & \textbf{Output Size} & \textbf{Input Size} & \textbf{Output Size} \\
\hline
%Layers & More table copy$^{\mathrm{a}}$& &  \\
\textbf{Conv$^{\mathrm{a}}$} & 224$\times$224$\times$3 & 112$\times$112$\times$16 & 224$\times$224$\times$3 & 112$\times$112$\times$32 & 224$\times$224$\times$3 & 112$\times$112$\times$32 \\
\cline{2-4} 
\hline
\textbf{Conv} & 112$\times$112$\times$16 & 56$\times$56$\times$28 & 112$\times$112$\times$32 & 56$\times$56$\times$56 & 112$\times$112$\times$32 & 56$\times$56$\times$64 \\ 
\cline{2-4} 
\hline
\textbf{Conv} & 56$\times$56$\times$28 & 28$\times$28$\times$64 & 56$\times$56$\times$56 & 28$\times$28$\times$256 & 56$\times$56$\times$64 & 28$\times$28$\times$256 \\
\cline{2-4} 
\hline
\textbf{D$^{\mathrm{b}}$} & 28$\times$28$\times$64 & 28$\times$28$\times$64 & 28$\times$28$\times$256 & 28$\times$28$\times$256 & 28$\times$28$\times$256 & 28$\times$28$\times$256 \\
\cline{2-4} 
\hline
\textbf{P$^{\mathrm{c}}$} & 28$\times$28$\times$64 & 28$\times$28$\times$256 & 28$\times$28$\times$256 & 28$\times$28$\times$256 & 28$\times$28$\times$256 & 28$\times$28$\times$256 \\
\cline{2-4} 
\hline
\textbf{D} & 28$\times$28$\times$256 & 14$\times$14$\times$256 & 28$\times$28$\times$256 & 14$\times$14$\times$256 & 28$\times$28$\times$256 & 14$\times$14$\times$256 \\ 
\cline{2-4} 
\hline
\textbf{P} & 14$\times$14$\times$256 & 14$\times$14$\times$256 & 14$\times$14$\times$256 & 14$\times$14$\times$512 & 14$\times$14$\times$256 & 14$\times$14$\times$512 \\
\cline{2-4} 
\hline
\textbf{D} & 14$\times$14$\times$256 & 14$\times$14$\times$256 & 14$\times$14$\times$512 & 14$\times$14$\times$512 & 14$\times$14$\times$512 & 14$\times$14$\times$512 \\ 
\cline{2-4} 
\hline
\textbf{P} & 14$\times$14$\times$256 & 14$\times$14$\times$256 & 14$\times$14$\times$512 & 14$\times$14$\times$512 & 14$\times$14$\times$512 & 14$\times$14$\times$512 \\
\cline{2-4} 
\hline

\textbf{D} & 14$\times$14$\times$256 & 14$\times$14$\times$256 & 14$\times$14$\times$512 & 14$\times$14$\times$512 & 14$\times$14$\times$512 & 14$\times$14$\times$512 \\
\cline{2-4} 
\hline
\textbf{P} & 14$\times$14$\times$256 & 14$\times$14$\times$256 & 14$\times$14$\times$512 & 14$\times$14$\times$512 & 14$\times$14$\times$512 & 14$\times$14$\times$512 \\
\cline{2-4} 
\hline
\textbf{D} & 14$\times$14$\times$256 & 14$\times$14$\times$256 & 14$\times$14$\times$512 & 14$\times$14$\times$512 & 14$\times$14$\times$512 & 14$\times$14$\times$512 \\ 
\cline{2-4} 
\hline
\textbf{P} & 14$\times$14$\times$256 & 14$\times$14$\times$256 & 14$\times$14$\times$512 & 14$\times$14$\times$512 & 14$\times$14$\times$512 & 14$\times$14$\times$512 \\
\cline{2-4} 
\hline
\textbf{D} & 14$\times$14$\times$256 & 14$\times$14$\times$256 & 14$\times$14$\times$512 & 14$\times$14$\times$512 & 14$\times$14$\times$512 & 14$\times$14$\times$512 \\ 
\cline{2-4} 
\hline
\textbf{P} & 14$\times$14$\times$256 & 14$\times$14$\times$256 & 14$\times$14$\times$512 & 14$\times$14$\times$512 & 14$\times$14$\times$512 & 14$\times$14$\times$512 \\
\cline{2-4} 
\hline
\textbf{D} & 14$\times$14$\times$256 & 14$\times$14$\times$256 & 14$\times$14$\times$512 & 14$\times$14$\times$512 & 14$\times$14$\times$512 & 14$\times$14$\times$512 \\
\cline{2-4} 
\hline
\textbf{P} & 14$\times$14$\times$256 & 14$\times$14$\times$256 & 14$\times$14$\times$512 & 14$\times$14$\times$512 & 14$\times$14$\times$512 & 14$\times$14$\times$512 \\

\cline{2-4} 
\hline
\textbf{D} & 14$\times$14$\times$256 & 7$\times$7$\times$256 & 14$\times$14$\times$512 & 7$\times$7$\times$512 & 14$\times$14$\times$512 & 7$\times$7$\times$512 \\ 
\cline{2-4} 
\hline
\textbf{P} & 7$\times$7$\times$256 & 7$\times$7$\times$1000 & 7$\times$7$\times$512 & 7$\times$7$\times$1000 & 7$\times$7$\times$512 & 7$\times$7$\times$1000 \\
\cline{2-4} 
\hline
\textbf{D} & 7$\times$7$\times$1000 & 7$\times$7$\times$1000 & 7$\times$7$\times$1000 & 7$\times$7$\times$1000 & 7$\times$7$\times$1000 & 7$\times$7$\times$1000 \\ 
\cline{2-4} 
\hline
\textbf{P} & 7$\times$7$\times$1000 & 7$\times$7$\times$1000 & 7$\times$7$\times$1000 & 7$\times$7$\times$1000 & 7$\times$7$\times$1000 & 7$\times$7$\times$1000 \\
\cline{2-4} 
\hline
\multicolumn{4}{l}{$^{\mathrm{a}}$Convolution layer.}\\
\multicolumn{4}{l}{$^{\mathrm{b}}$Depthwise convolution layer.}\\
\multicolumn{4}{l}{$^{\mathrm{c}}$Pointwise convolution layer.}
\end{tabular}
\label{tab1}
\end{center}
\end{table}

%\subsection{Resource Identification Initiative}
%To take part in the Resource Identification Initiative, please use the corresponding catalog number and RRID in your current manuscript. For more information about the project and for steps on how to search for an RRID, please click \href{http://www.frontiersin.org/files/pdf/letter_to_author.pdf}{here}.

%\subsection{Life Science Identifiers}
%Life Science Identifiers (LSIDs) for ZOOBANK registered names or nomenclatural acts should be listed in the manuscript before the keywords. For more information on LSIDs please see \href{http://www.frontiersin.org/about/AuthorGuidelines#InclusionofZoologicalNomenclature}{Inclusion of Zoological Nomenclature} section of the guidelines.

\section{Result}
\label{result}

In this section, we design three sets of experiments to investigate how variations in the proposed architecture affect classification accuracies on the IMAGENET dataset \cite{imagenet}. Note that all the different neural network models considered in this section is illustrated in the supplementary material. The first set of experiment investigates the performance of the proposed architecture with and without pooling layers and fully connected layers. The proposed architecture as in table \ref{tab1} for the core size of 1024$\times$1024 (we refer this as base model in the entire manuscript) is trained on the IMAGENET dataset with and without pooling layers, similarly with and without fully connected layers. Network without pooling layer is exactly same as the mentioned architecture in table \ref{tab1}, but for network with pooling layer, all the layers with stride of 2 is inserted with pooling layers. Fully connected layer of size 1024$\times$1000 is added at the end of the proposed architecture in network with fully connected layer. From table \ref{tab2}, it can be seen that there is no significant improvement in classification accuracies with and without pooling layers and fully connected layers. Hence, we have completely removed the pooling layers and fully connected layers from the proposed architecture.

The proposed architecture is trained on IMAGENET dataset with batch normalization technique before ReLU activation function in every layer. We would consider binary activations in future work, as the purpose of this work is to propose a novel CNN architecture that is neuromorphic hardware-friendly. Hence, how quantized activations affect classification accuracies is beyond the scope of our current work. The network does not converge without batch normalization. For the next set of experiments, there are three architectures for three different core sizes as mentioned in table \ref{tab1}. IMAGENET dataset is trained on all the three different architectures. From table \ref{tab3}, it can be seen that the classification accuracy for different architectures improve with bigger neuromorphic core sizes, as larger network architectures can be mapped onto larger core sizes.

The third set of experiment involves addition of layers on top of the base model as proposed in table \ref{tab1} for the core size of 1024$\times$1024. Here, we have considered adding three layers (depthwise and pointwise convolution layer, standard convolution layer and fully connected layers) separately on to the base model to test the accuracy on IMAGENET dataset. For the addition of depthwise and pointwise convolution layers (1 DP layer as in table \ref{tab4}), we have added one depthwise and one pointwise convolution respectively to the end of the base model and trained the network on IMAGENET dataset. For the addition of a standard convolution layer (1 Conv layer as in table \ref{tab4}), we have added a convolution layer to the front of the base model. Similarly, for the addition of fully connected layer (2 FC layer as in table \ref{tab4}), we have added 2 fully connected layer at the end of the base model. For this, we changed the output size of the last pointwise convolution layer to $7\times7\times1024$, instead of $7\times7\times1000$. Note that adding fully connected layer will increase fan-in degree and will not fit onto 1024 core, i.e. fully connected layer is not a hardware-friendly layer. 1 DP-1 Conv layer as in table \ref{tab4} is the addition of both depthwise and pointwise convolution layer at the end of the base model along with the addition of convolution layer at the front of the base model. Table \ref{tab4} shows the results for addition of layers to the base model. It can be seen that all the accuracies are more than the accuracy of the base model which is 68.14$\%$ as given in table \ref{tab2}. Adding a standard convolution layer at the front of the base model gives better result than adding a depthwise and pointwise convolution layer at the end of the base model. Whereas, adding two fully connected layers at the end of the base model does not show much improvement in accuracy as in the aforementioned case of addition of single layers. But, addition of both depthwise-pointwise convolution layer along with standard convolution layer shows the best result among the four, which is around the same accuracy claimed by VGGNET.     

\begin{table}[htbp]
\caption{With and without pooling and fully connected layers}
\begin{center}
\begin{tabular}{|c|c|c|}
\hline
\multicolumn{3}{|c|}{\textbf{Classification Accuracy (\%)}} \\
\cline{1-3}
\textbf{\textit{Base model$^{\mathrm{*}}$}}&\textbf{\textit{With pooling layer}}&\textbf{\textit{With FC$^{\mathrm{**}}$ layer}} \\
\hline
68.14 & 68.85 & 68.21  \\
\hline
\multicolumn{3}{l}{$^{\mathrm{*}}$The proposed architecture as in table \ref{tab1} for the core size of 1024$\times$1024.}\\
\multicolumn{3}{l}{$^{\mathrm{**}}$Fully connected.}
\end{tabular}
\label{tab2}
\end{center}
\end{table}

\begin{table}[htbp]
\caption{Using different architectures for different core sizes}
\begin{center}
\begin{tabular}{|c|c|c|c|}
\hline
\multicolumn{3}{|c|}{\textbf{Classification Accuracy (\%)}} \\
\cline{1-3}
\textbf{\textit{256$\times$256}}&\textbf{\textit{512$\times$512}}&\textbf{\textit{1024$\times$1024}} \\
\hline
62.46 & 67.78 & 68.14  \\
\hline 
%\multicolumn{4}{l}{$^{\mathrm{a}}$Sample of a Table footnote.}
\end{tabular}
\label{tab3}
\end{center}
\end{table}

\begin{table}[htbp]
\caption{Addition of layers to base model}
\begin{center}
\begin{tabular}{|c|c|c|c|}
\hline
\multicolumn{4}{|c|}{\textbf{Classification Accuracy (\%)}} \\
\cline{1-4}
\textbf{\textit{1 DP$^{\mathrm{*}}$ layer}}&\textbf{\textit{1 Conv layer}}&\textbf{\textit{1 DP-1 Conv layer}}&\textbf{\textit{2 FC layer}} \\
\hline
68.73 & 69.43 & 70.186 & 68.21 \\
\hline 
\multicolumn{4}{l}{$^{\mathrm{*}}$Depthwise and Pointwise convolution.}
\end{tabular}
\label{tab4}
\end{center}
\end{table}

%For additional requirements for specific article types and further information please refer to \href{http://www.frontiersin.org/about/AuthorGuidelines#AdditionalRequirements}{Author Guidelines}.

\section{Conclusion}
\label{Conc}

Neuromorphic hardware friendly neural networks are customized for a specific neuromorphic hardware such that it can then be easily mapped onto the hardware. In our work, the proposed neuromorphic hardware friendly CNN is compatible with a neuromorphic hardware with crossbar array of synapses in a neuromorphic core. One of the motivation of the proposed architecture is to maximise the utilization of the crossbar architecture, which may not be possible with existing CNNs, but we can then modify to fit onto the hardware. By doing so, we avoid splitting the weight matrix of a particular neuron into more than one core during mapping. Splitting requires intermediate neurons which increase the hardware overhead and also effectively introduces new non-linearity into the neural network which affects accuracies. Also mapping of the existing CNNs onto the neuromorphic chip requires more than one neuromorphic chip with limited cores. The deeper layers in the existing CNNs with bigger feature maps also require splitting the weight matrix. This splitting of weight matrix is completely removed in our proposed hardware friendly architecture.    

Different architectures for different neuromorphic core sizes in the results further shows that the architecture can be tailored for different core sizes. It also shows that larger the core size, the larger the network, and the better the classification accuracy. Chip design however limits the size of the neuromorphic cores. We have proposed a novel architecture without pooling and fully connected layers. %(cys: BN has bias). 
The results in the section \ref{result} further justifies not using the aforementioned layers as classification accuracies are not affected.

In our current study, we only studied CNNs that do not have connections skipping layers. Hence, residual networks \cite{ResidualNET} are not considered. Skipped connections would increase the fan-in/fan-out degree of the neurons in a neuromorphic chip. We would consider mapping of such networks in a future work. %(cys: i have included this here. i dont think skipped connection cannot be implemented in a neuromorphic h/w other than increasing the fan-in/fan-out degree, please verify)
We would also consider other hardware constraints such as low precision of weights and binary activations in the future. For Resistive Random Access Memory (RRAM) devices, synaptic noise and variability will have to be considered as well.

%(cys: the below paragraph is not necessary, please remove)
%The crossbar architecture of synapses in a neuromorphic core mentioned here is based on the Resistive Random Access Memories (RRAM) []. Apart from RRAM, one can also explore the use of CMOS devices such as the floating gate MOSFETs \cite{Roshan_DSTDP_TNNLS, Roshan_TSTDP_TNNLS} or nano-technology devices such as the Phase Change Memories (PCM) [] and Spin-Transfer Torque Magnetic Random Access Memories (STT-MRAMs) [] as a synaptic device \cite{Roshan_DSTDP_IJCNN, Roshan_TSTDP_ISCAS}. Device modelling and noise characterizations of these devices become the first step to realise its usage as a synaptic device. The research in this field is progressing drastically with invention of new materials for synaptic device to implement several types of synaptic plasticity \cite{Rohit}. 

%\section*{Conflict of Interest Statement}
%All financial, commercial or other relationships that might be perceived by the academic community as representing a potential conflict of interest must be disclosed. If no such relationship exists, authors will be asked to confirm the following statement: 

%The authors declare that the research was conducted in the absence of any commercial or financial relationships that could be construed as a potential conflict of interest.

\section*{Author Contributions}

The first and second author designed and proposed the architecture and experimental framework. The first author wrote the manuscript; the second author edited the manuscript and conducted some experiments. The third author is involved in generating the figures and conducting some experiments.   

%The Author Contributions section is mandatory for all articles, including articles by sole authors. If an appropriate statement is not provided on submission, a standard one will be inserted during the production process. The Author Contributions statement must describe the contributions of individual authors referred to by their initials and, in doing so, all authors agree to be accountable for the content of the work. Please see  \href{http://home.frontiersin.org/about/author-guidelines#AuthorandContributors}{here} for full authorship criteria.

\section*{Funding}
This research is supported by Programmatic grant no. A1687b0033 from the Singapore governments Research, Innovation and Enterprise 2020 plan (Advanced Manufacturing and Engineering domain).

\section*{Acknowledgments}
Authors would like to thank Assoc Prof Arindam Basu
from Nanyang Technological University, Singapore for useful
discussions.
%This is a short text to acknowledge the contributions of specific colleagues, institutions, or agencies that aided the efforts of the authors.

%\section*{Supplemental Data}
% \href{http://home.frontiersin.org/about/author-guidelines#SupplementaryMaterial}{Supplementary Material} should be uploaded separately on submission, if there are Supplementary Figures, please include the caption in the same file as the figure. LaTeX Supplementary Material templates can be found in the Frontiers LaTeX folder.

%\section*{Data Availability Statement}
%The datasets [GENERATED/ANALYZED] for this study can be found in the [NAME OF REPOSITORY] [LINK].
% Please see the availability of data guidelines for more information, at https://www.frontiersin.org/about/author-guidelines#AvailabilityofData

\bibliographystyle{frontiersinSCNS_ENG_HUMS} % for Science, Engineering and Humanities and Social Sciences articles, for Humanities and Social Sciences articles please include page numbers in the in-text citations
\bibliography{test}

\begin{thebibliography}{33}
\providecommand{\natexlab}[1]{#1}
\expandafter\ifx\csname urlstyle\endcsname\relax
  \providecommand{\doi}[1]{doi:\discretionary{}{}{}#1}\else
  \providecommand{\doi}{doi:\discretionary{}{}{}\begingroup
  \urlstyle{rm}\Url}\fi
\providecommand{\selectlanguage}[1]{\relax}
\providecommand{\bibAnnoteFile}[1]{%
  \IfFileExists{#1}{\begin{quotation}\noindent\textsc{Key:} #1\\
  \textsc{Annotation:}\ \input{#1}\end{quotation}}{}}
\providecommand{\bibAnnote}[2]{%
  \begin{quotation}\noindent\textsc{Key:} #1\\
  \textsc{Annotation:}\ #2\end{quotation}}

\bibitem[{{Akopyan} et~al.(2015){Akopyan}, {Sawada}, {Cassidy},
  {Alvarez-Icaza}, {Arthur}, {Merolla} et~al.}]{truenorth_TCAD}
{Akopyan}, F., {Sawada}, J., {Cassidy}, A., {Alvarez-Icaza}, R., {Arthur}, J.,
  {Merolla}, P., et~al. (2015).
\newblock Truenorth: Design and tool flow of a 65 mw 1 million neuron
  programmable neurosynaptic chip.
\newblock \emph{IEEE Transactions on Computer-Aided Design of Integrated
  Circuits and Systems} 34, 1537--1557
\bibAnnoteFile{truenorth_TCAD}

\bibitem[{{Ambrogio} et~al.(2014{\natexlab{a}}){Ambrogio}, {Balatti}, {Cubeta},
  {Calderoni}, {Ramaswamy}, and {Ielmini}}]{RRAM_variability}
{Ambrogio}, S., {Balatti}, S., {Cubeta}, A., {Calderoni}, A., {Ramaswamy}, N.,
  and {Ielmini}, D. (2014{\natexlab{a}}).
\newblock Statistical fluctuations in hfox resistive-switching memory: Part i -
  set/reset variability.
\newblock \emph{IEEE Transactions on Electron Devices} 61, 2912--2919.
\newblock \doi{10.1109/TED.2014.2330200}
\bibAnnoteFile{RRAM_variability}

\bibitem[{{Ambrogio} et~al.(2014{\natexlab{b}}){Ambrogio}, {Balatti}, {Cubeta},
  {Calderoni}, {Ramaswamy}, and {Ielmini}}]{RRAM_RTN}
{Ambrogio}, S., {Balatti}, S., {Cubeta}, A., {Calderoni}, A., {Ramaswamy}, N.,
  and {Ielmini}, D. (2014{\natexlab{b}}).
\newblock Statistical fluctuations in hfox resistive-switching memory: Part ii
  - random telegraph noise.
\newblock \emph{IEEE Transactions on Electron Devices} 61, 2912--2919.
\newblock \doi{10.1109/TED.2014.2330200}
\bibAnnoteFile{RRAM_RTN}

\bibitem[{{Amir} et~al.(2013){Amir}, {Datta}, {Risk}, {Cassidy}, {Kusnitz},
  {Esser} et~al.}]{IBM_corelet}
{Amir}, A., {Datta}, P., {Risk}, W.~P., {Cassidy}, A.~S., {Kusnitz}, J.~A.,
  {Esser}, S.~K., et~al. (2013).
\newblock Cognitive computing programming paradigm: A corelet language for
  composing networks of neurosynaptic cores.
\newblock In \emph{The 2013 International Joint Conference on Neural Networks
  (IJCNN)}. 1--10
\bibAnnoteFile{IBM_corelet}

\bibitem[{Appuswamy et~al.(2016)Appuswamy, Nayak, Arthur, Esser, Merolla,
  McKinstry et~al.}]{toeplitz_IBM}
Appuswamy, R., Nayak, T.~K., Arthur, J.~V., Esser, S.~K., Merolla, P.,
  McKinstry, J.~L., et~al. (2016).
\newblock Structured convolution matrices for energy-efficient deep learning
\bibAnnoteFile{toeplitz_IBM}

\bibitem[{Clevert et~al.(2015)Clevert, Unterthiner, and Hochreiter}]{ELU}
Clevert, D., Unterthiner, T., and Hochreiter, S. (2015).
\newblock Fast and accurate deep network learning by exponential linear units
  (elus).
\newblock \emph{CoRR} abs/1511.07289
\bibAnnoteFile{ELU}

\bibitem[{Courbariaux et~al.(2016)Courbariaux, Hubara, Soudry, El-Yaniv, and
  Bengio}]{BNN}
Courbariaux, M., Hubara, I., Soudry, D., El-Yaniv, R., and Bengio, Y. (2016).
\newblock Binarized neural networks: Training deep neural networks with weights
  and activations constrained to+ 1 or-1.
\newblock \emph{arXiv preprint arXiv:1602.02830}
\bibAnnoteFile{BNN}

\bibitem[{Deng et~al.(2009)Deng, Dong, Socher, Li, Li, and Fei-Fei}]{imagenet}
Deng, J., Dong, W., Socher, R., Li, L.-J., Li, K., and Fei-Fei, L. (2009).
\newblock {ImageNet: A Large-Scale Hierarchical Image Database}.
\newblock In \emph{CVPR09}
\bibAnnoteFile{imagenet}

\bibitem[{Esser et~al.(2016)Esser, Merolla, Arthur, Cassidy, Appuswamy,
  Andreopoulos et~al.}]{IBM_Esser}
Esser, S.~K., Merolla, P.~A., Arthur, J.~V., Cassidy, A.~S., Appuswamy, R.,
  Andreopoulos, A., et~al. (2016).
\newblock Convolutional networks for fast, energy-efficient neuromorphic
  computing.
\newblock \emph{Proceedings of the National Academy of Sciences} 113,
  11441--11446.
\newblock \doi{10.1073/pnas.1604850113}
\bibAnnoteFile{IBM_Esser}

\bibitem[{Gopalakrishnan(2019)}]{Roshan_RRAM_ARXIV}
Gopalakrishnan, R. (2019).
\newblock Rram based neuromorphic algorithms
\bibAnnoteFile{Roshan_RRAM_ARXIV}

\bibitem[{Gopalakrishnan et~al.(2019)Gopalakrishnan, Kumar, and
  Chua}]{MAD_Arxiv}
Gopalakrishnan, R., Kumar, A. J.~S., and Chua, Y. (2019).
\newblock Mad: Mapping and debugging framework for implementing deep neural
  network onto a neuromorphic chip with crossbar array of synapses
\bibAnnoteFile{MAD_Arxiv}

\bibitem[{Graham(2014)}]{Fractional_max}
Graham, B. (2014).
\newblock Fractional max-pooling.
\newblock \emph{CoRR} abs/1412.6071
\bibAnnoteFile{Fractional_max}

\bibitem[{Gray(2006)}]{toeplitz}
Gray, R.~M. (2006).
\newblock Toeplitz and circulant matrices: A review
\bibAnnoteFile{toeplitz}

\bibitem[{He et~al.(2016)He, Zhang, Ren, and Sun}]{ResidualNET}
He, K., Zhang, X., Ren, S., and Sun, J. (2016).
\newblock Deep residual learning for image recognition.
\newblock In \emph{Proceedings of the IEEE conference on computer vision and
  pattern recognition}. 770--778
\bibAnnoteFile{ResidualNET}

\bibitem[{Howard et~al.(2017)Howard, Zhu, Chen, Kalenichenko, Wang, Weyand
  et~al.}]{MobileNet}
Howard, A.~G., Zhu, M., Chen, B., Kalenichenko, D., Wang, W., Weyand, T.,
  et~al. (2017).
\newblock Mobilenets: Efficient convolutional neural networks for mobile vision
  applications.
\newblock \emph{arXiv preprint arXiv:1704.04861}
\bibAnnoteFile{MobileNet}

\bibitem[{Hu et~al.(2016)Hu, Strachan, Li, Grafals, Davila, Graves
  et~al.}]{MVM}
Hu, M., Strachan, J.~P., Li, Z., Grafals, E.~M., Davila, N., Graves, C., et~al.
  (2016).
\newblock Dot-product engine for neuromorphic computing: Programming 1t1m
  crossbar to accelerate matrix-vector multiplication.
\newblock In \emph{2016 53nd ACM/EDAC/IEEE Design Automation Conference (DAC)}.
  1--6
\bibAnnoteFile{MVM}

\bibitem[{Iandola et~al.(2016)Iandola, Han, Moskewicz, Ashraf, Dally, and
  Keutzer}]{SqueezeNet}
Iandola, F.~N., Han, S., Moskewicz, M.~W., Ashraf, K., Dally, W.~J., and
  Keutzer, K. (2016).
\newblock Squeezenet: Alexnet-level accuracy with 50x fewer parameters and< 0.5
  mb model size.
\newblock \emph{arXiv preprint arXiv:1602.07360}
\bibAnnoteFile{SqueezeNet}

\bibitem[{Ji et~al.(2018)Ji, Zhang, Chen, and Xie}]{NC_constraints}
Ji, Y., Zhang, Y., Chen, W., and Xie, Y. (2018).
\newblock Bridge the gap between neural networks and neuromorphic hardware with
  a neural network compiler.
\newblock In \emph{Proceedings of the Twenty-Third International Conference on
  Architectural Support for Programming Languages and Operating Systems} (ACM),
  448--460
\bibAnnoteFile{NC_constraints}

\bibitem[{{Ji} et~al.(2016){Ji}, {Zhang}, {Li}, {Chi}, {Jiang}, {Qu}
  et~al.}]{neutrams}
{Ji}, Y., {Zhang}, Y., {Li}, S., {Chi}, P., {Jiang}, C., {Qu}, P., et~al.
  (2016).
\newblock Neutrams: Neural network transformation and co-design under
  neuromorphic hardware constraints.
\newblock In \emph{2016 49th Annual IEEE/ACM International Symposium on
  Microarchitecture (MICRO)}. 1--13
\bibAnnoteFile{neutrams}

\bibitem[{Krizhevsky et~al.(2012)Krizhevsky, Sutskever, and Hinton}]{AlexNet}
Krizhevsky, A., Sutskever, I., and Hinton, G.~E. (2012).
\newblock Imagenet classification with deep convolutional neural networks.
\newblock In \emph{Advances in neural information processing systems}.
  1097--1105
\bibAnnoteFile{AlexNet}

\bibitem[{Lee et~al.(2018)Lee, Cui, Somu, Luo, Zhou, Tang et~al.}]{neurosim}
Lee, M., Cui, Y., Somu, T., Luo, T., Zhou, J., Tang, W., et~al. (2018).
\newblock A system-level simulator for rram-based neuromorphic computing chips
\bibAnnoteFile{neurosim}

\bibitem[{Lin et~al.(2013)Lin, Chen, and Yan}]{NIN}
Lin, M., Chen, Q., and Yan, S. (2013).
\newblock Network in network.
\newblock \emph{CoRR} abs/1312.4400
\bibAnnoteFile{NIN}

\bibitem[{P et~al.(2009)P, Daniel, Jochen, Jens, Eilif, Dejan et~al.}]{PyNN}
P, D.~A., Daniel, B., Jochen, E., Jens, K., Eilif, M., Dejan, P., et~al.
  (2009).
\newblock Pynn: A common interface for neuronal network simulators 2, 11.
\newblock \doi{10.3389/neuro.11.011.2008}
\bibAnnoteFile{PyNN}

\bibitem[{Prezioso et~al.(2015)Prezioso, Merrikh-Bayat, Hoskins, Adam,
  Likharev, and Strukov}]{crossbar}
Prezioso, M., Merrikh-Bayat, F., Hoskins, B.~D., Adam, G.~C., Likharev, K.~K.,
  and Strukov, D.~B. (2015).
\newblock Training and operation of an integrated neuromorphic network based on
  metal-oxide memristors 521, 61--64
\bibAnnoteFile{crossbar}

\bibitem[{Rastegari et~al.(2016)Rastegari, Ordonez, Redmon, and
  Farhadi}]{XNORNET}
Rastegari, M., Ordonez, V., Redmon, J., and Farhadi, A. (2016).
\newblock Xnor-net: Imagenet classification using binary convolutional neural
  networks.
\newblock In \emph{European Conference on Computer Vision} (Springer), 525--542
\bibAnnoteFile{XNORNET}

\bibitem[{Simonyan and Zisserman(2014)}]{VGGNet}
Simonyan, K. and Zisserman, A. (2014).
\newblock Very deep convolutional networks for large-scale image recognition.
\newblock \emph{arXiv preprint arXiv:1409.1556}
\bibAnnoteFile{VGGNet}

\bibitem[{Springenberg et~al.(2014)Springenberg, Dosovitskiy, Brox, and
  Riedmiller}]{without_pool}
Springenberg, J.~T., Dosovitskiy, A., Brox, T., and Riedmiller, M.~A. (2014).
\newblock Striving for simplicity: The all convolutional net.
\newblock \emph{CoRR} abs/1412.6806
\bibAnnoteFile{without_pool}

\bibitem[{Szegedy et~al.(2015)Szegedy, Liu, Jia, Sermanet, Reed, Anguelov
  et~al.}]{GoogLeNet}
Szegedy, C., Liu, W., Jia, Y., Sermanet, P., Reed, S., Anguelov, D., et~al.
  (2015).
\newblock Going deeper with convolutions.
\newblock In \emph{Computer Vision and Pattern Recognition (CVPR), 2015 IEEE
  Conference on} (IEEE), 1--9
\bibAnnoteFile{GoogLeNet}

\bibitem[{{Voelker} et~al.(2017){Voelker}, {Benjamin}, {Stewart}, {Boahen}, and
  {Eliasmith}}]{NEF}
{Voelker}, A.~R., {Benjamin}, B.~V., {Stewart}, T.~C., {Boahen}, K., and
  {Eliasmith}, C. (2017).
\newblock Extending the neural engineering framework for nonideal silicon
  synapses.
\newblock In \emph{2017 IEEE International Symposium on Circuits and Systems
  (ISCAS)}. 1--4.
\newblock \doi{10.1109/ISCAS.2017.8050810}
\bibAnnoteFile{NEF}

\bibitem[{Wan et~al.(2013)Wan, Zeiler, Zhang, Cun, and Fergus}]{DropConnect}
Wan, L., Zeiler, M., Zhang, S., Cun, Y.~L., and Fergus, R. (2013).
\newblock Regularization of neural networks using dropconnect.
\newblock In \emph{Proceedings of the 30th International Conference on Machine
  Learning}, eds. S.~Dasgupta and D.~McAllester (Atlanta, Georgia, USA: PMLR),
  vol.~28 of \emph{Proceedings of Machine Learning Research}, 1058--1066
\bibAnnoteFile{DropConnect}

\bibitem[{Yakopcic et~al.(2016)Yakopcic, Alom, and Taha}]{Alom_2016}
Yakopcic, C., Alom, M.~Z., and Taha, T.~M. (2016).
\newblock Memristor crossbar deep network implementation based on a
  convolutional neural network.
\newblock In \emph{2016 International Joint Conference on Neural Networks
  (IJCNN)}. 963--970
\bibAnnoteFile{Alom_2016}

\bibitem[{Yakopcic et~al.(2017)Yakopcic, Alom, and Taha}]{Alom_2017}
Yakopcic, C., Alom, M.~Z., and Taha, T.~M. (2017).
\newblock Extremely parallel memristor crossbar architecture for convolutional
  neural network implementation.
\newblock In \emph{2017 International Joint Conference on Neural Networks
  (IJCNN)}. 1696--1703
\bibAnnoteFile{Alom_2017}

\bibitem[{Zhou et~al.(2016)Zhou, Wu, Ni, Zhou, Wen, and Zou}]{DOREFA}
Zhou, S., Wu, Y., Ni, Z., Zhou, X., Wen, H., and Zou, Y. (2016).
\newblock Dorefa-net: Training low bitwidth convolutional neural networks with
  low bitwidth gradients.
\newblock \emph{arXiv preprint arXiv:1606.06160}
\bibAnnoteFile{DOREFA}

\end{thebibliography}

\end{document}